\documentclass[10pt,twocolumn,letterpaper]{article}

\usepackage[pagenumbers]{cvpr}      % To produce the REVIEW version
\definecolor{cvprblue}{rgb}{0.21,0.49,0.74}
\usepackage[pagebackref,breaklinks,colorlinks,allcolors=cvprblue]{hyperref}

\def\paperID{5794} % *** Enter the Paper ID here
\def\confName{CVPR}
\def\confYear{2026}

\title{Neural Electromagnetic Fields for High-Resolution Material Parameter Reconstruction}

\author{
Zhe Chen$^{1}$ \quad
Peilin Zheng$^{1}$ \quad
Wenshuo Chen$^{2}$ \quad
Xiucheng Wang$^{1}$ \quad
Yutao Yue$^{2}$ \quad
Nan Cheng$^{1}$\\[0.2em]
$^{1}$Xidian University \quad
$^{2}$The Hong Kong University of Science and Technology
}

\begin{document}
\twocolumn[{%
\renewcommand\twocolumn[1][]{#1}%
\maketitle
\begin{center}
    \centering
    \captionsetup{type=figure}
    \includegraphics[width=\textwidth,height=8.5cm]{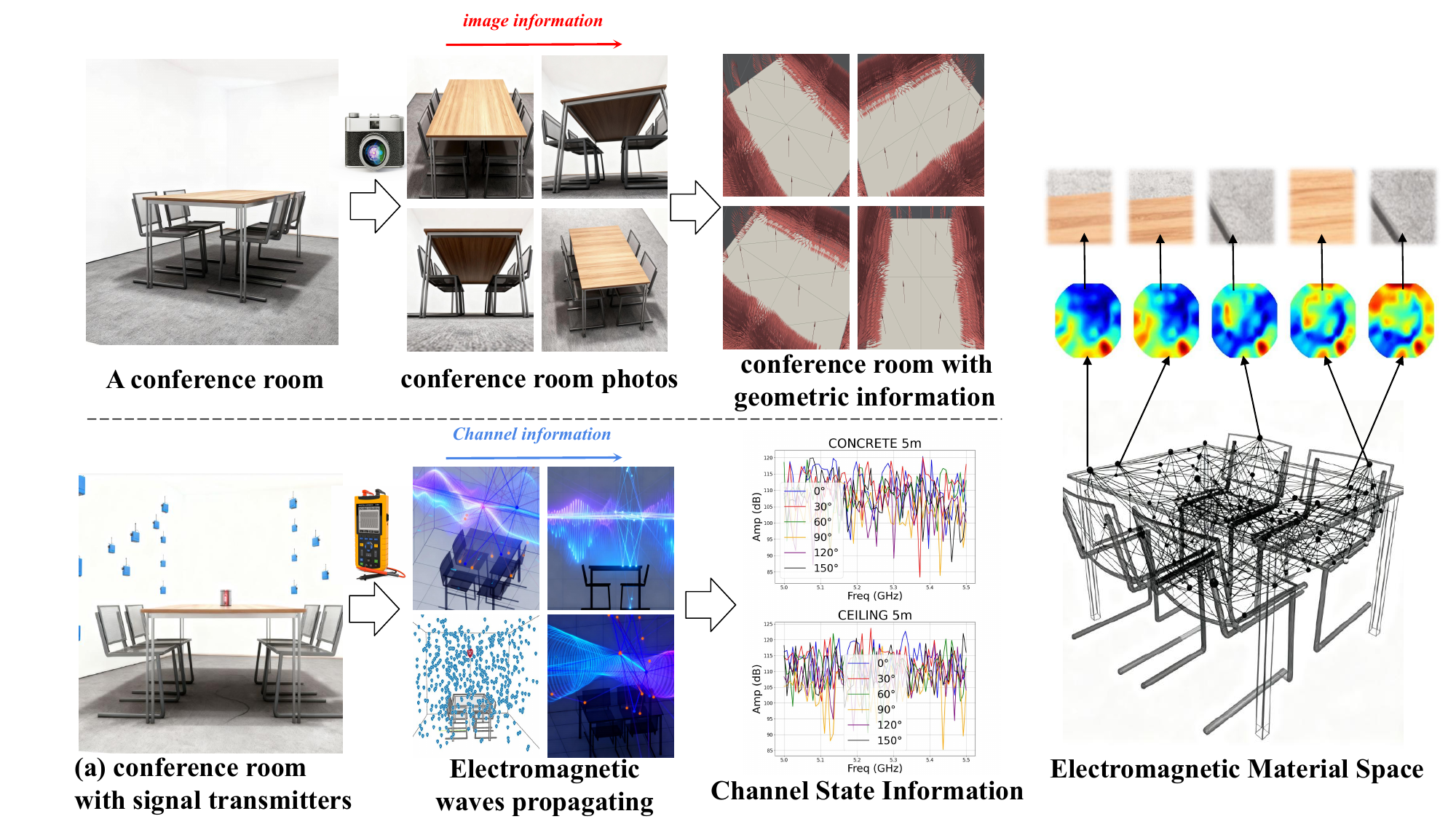}
    \captionof{figure}{Overview of NEMF framework showing multi-modal fusion for functional digital twins.}
\end{center}%
}]
\maketitle
\begin{abstract} Creating functional Digital Twins, simulatable 3D replicas of the real world, is a central challenge in computer vision. Current methods like NeRF produce visually rich but functionally incomplete twins. The key barrier is the lack of underlying material properties (e.g., permittivity, conductivity). Acquiring this information for every point in a scene via non-contact, non-invasive sensing is a primary goal, but it demands solving a notoriously ill-posed physical inversion problem. Standard remote signals, like images and radio frequencies (RF), deeply entangle the unknown geometry, ambient field, and target materials.
We introduce NEMF, a novel framework for dense, non-invasive physical inversion designed to build functional digital twins. Our key insight is a systematic disentanglement strategy. NEMF leverages high-fidelity geometry from images as a powerful anchor, which first enables the resolution of the ambient field. By constraining both geometry and field using only non-invasive data, the original ill-posed problem transforms into a well-posed, physics-supervised learning task.
This transformation unlocks our core inversion module: a decoder. Guided by ambient RF signals and a differentiable layer incorporating physical reflection models, it learns to explicitly output a continuous, spatially-varying field of the scene's underlying material parameters. We validate our framework on high-fidelity synthetic datasets. Experiments show our non-invasive inversion reconstructs these material maps with high accuracy, and the resulting functional twin enables high-fidelity physical simulation. This advance moves beyond passive visual replicas, enabling the creation of truly functional and simulatable models of the physical world. \end{abstract} 
\section{Introduction}
\label{sec:intro}

\begin{figure*}[t]
  \centering
  \subfloat[NeRF visual twin (non-functional)]{\includegraphics[width=0.3\textwidth]{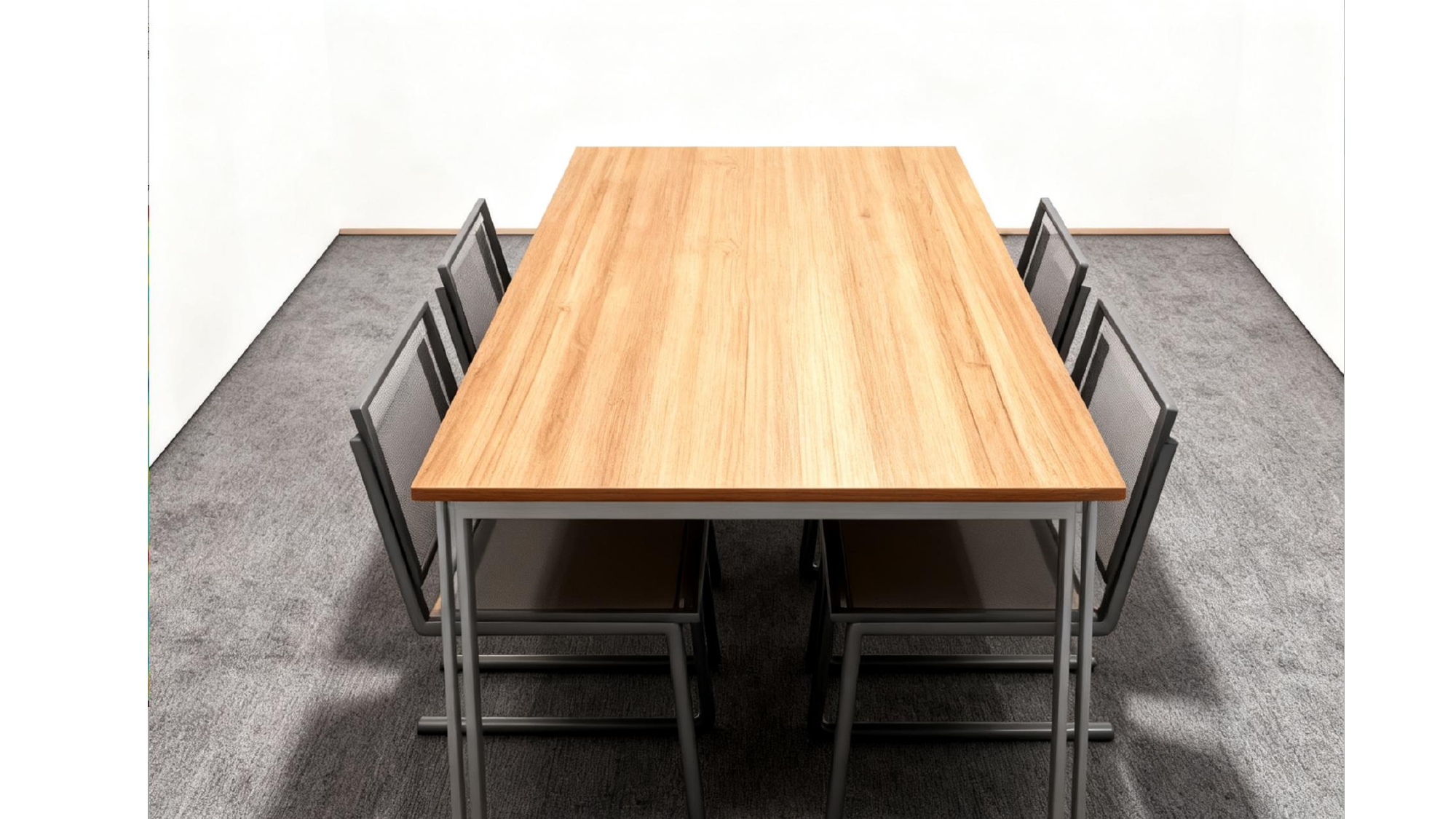}} 
  \subfloat[Ill-posed EM inversion (no geometry)]{\includegraphics[width=0.3\textwidth]{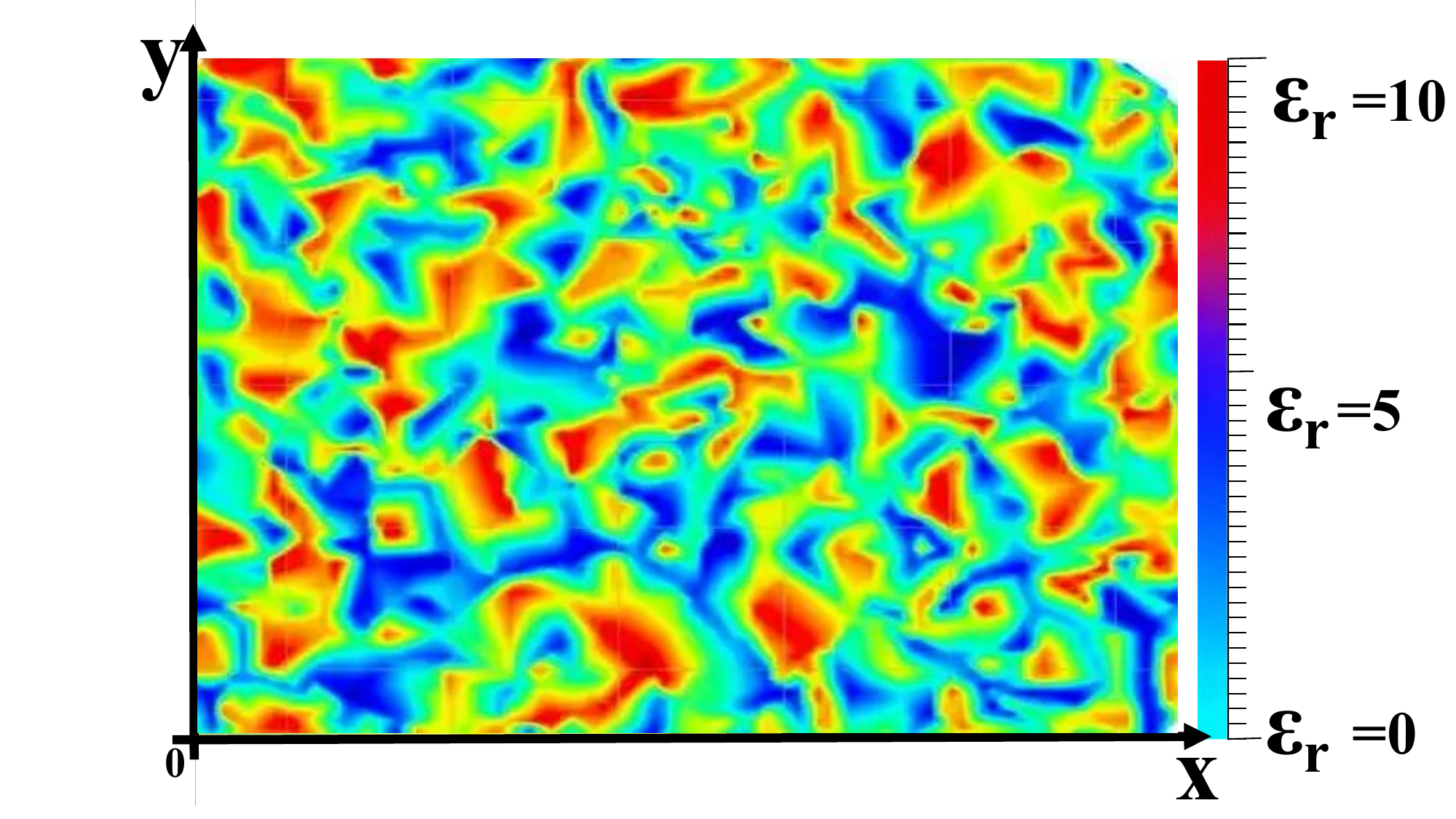}} 
  \subfloat[NEMF (Ours): A Functional Twin]{\includegraphics[width=0.3\textwidth]{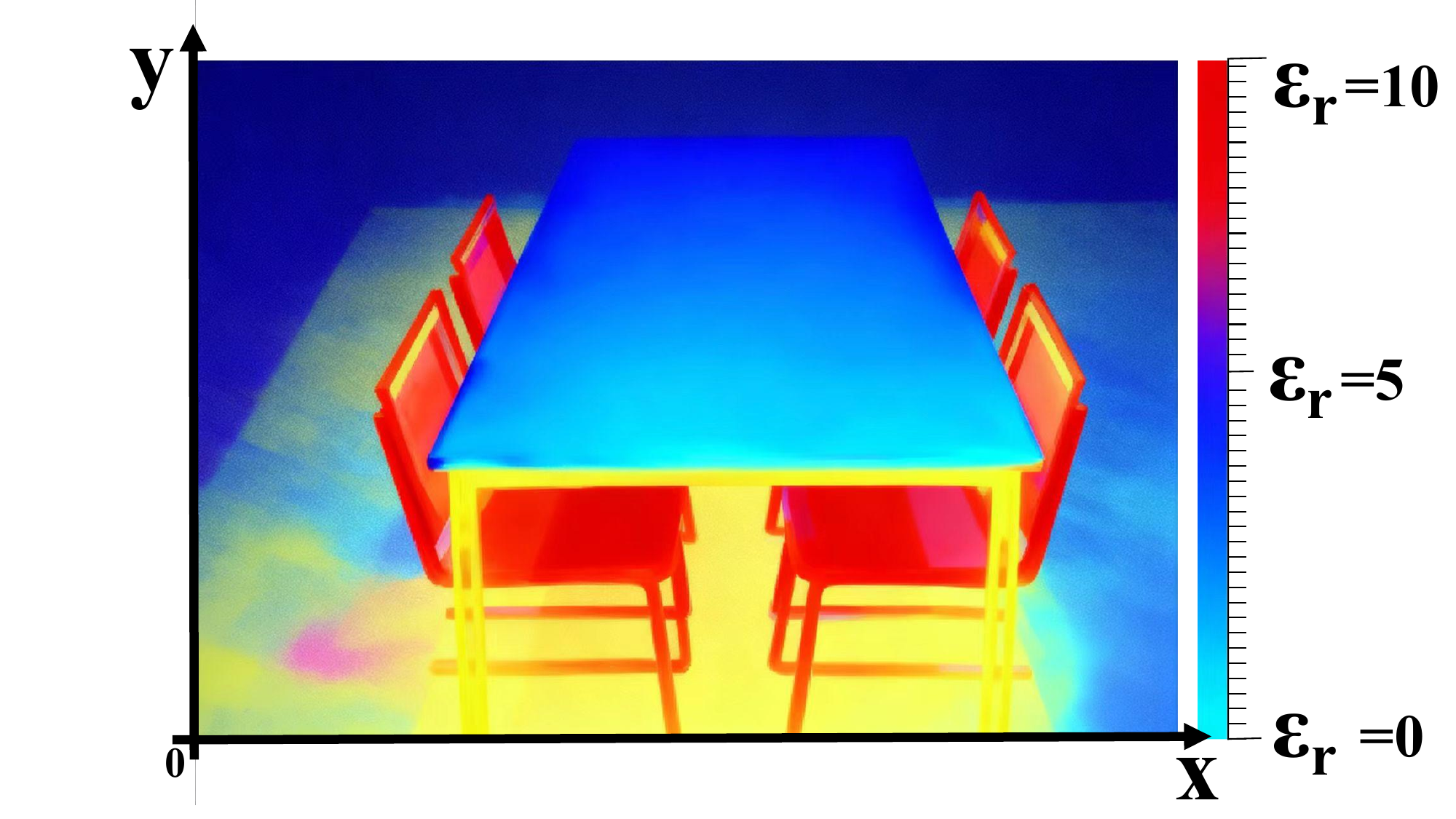}}
  \caption{Motivation. (a) Standard Neural Radiance Fields (NeRF) create photorealistic \textit{visual} twins, but their parameters are implicit and non-physical. (b) Inverting electromagnetic (EM) parameters from sparse signals alone is a highly ill-posed problem. (c) Our work, NEMF, fuses both modalities (visual geometry + physical signals) to create a \textit{functional} digital twin with explicit, simulatable material properties.}
  \label{fig:motivation}
\end{figure*}

The reconstruction of high-fidelity, interactive three-dimensional "Digital Twins" from multi-view images is a central pursuit in modern computer vision and graphics. This field has recently been revolutionized by Neural Radiance Fields (NeRF)~\cite{nerf} and its many variants~\cite{barron2021mipnerf, muller2022instant, kerbl2023gaussians, verbin2022ref}. NeRF excels at learning a continuous volumetric representation from two-dimensional images, enabling the synthesis of photorealistic \emph{visual twins} with unprecedented fidelity. This has provided powerful tools for view synthesis, scene reconstruction, and augmented reality.

While these representations are visually striking, they are \textbf{not "functional" physical models}. The parameters of a NeRF are learned implicitly from data to reproduce optical effects; they are not explicitly constrained by physical laws or mapped to tangible material properties. For example, a NeRF can render the appearance of a wall perfectly, yet its internal weights cannot be used to simulate how that wall blocks or reflects electromagnetic waves. This fundamental gap between \textit{visual appearance} and \textit{physical function} hinders applications such as predicting indoor WiFi coverage for augmented reality or enabling non-line-of-sight (NLOS) sensing in robotics.

To create truly "functional" digital twins, we must integrate underlying \textbf{physical properties} into these three-dimensional scenes. This paper addresses this challenge by focusing on \textbf{electromagnetic (EM) properties} (e.g., permittivity \(\epsilon_r\) and conductivity \(\sigma\)). The only scalable, non-invasive method to probe these properties is by using ubiquitous radio frequency (RF) signals, specifically Channel State Information (CSI). \textbf{However, a critical gap exists:} inverting physical parameters from CSI alone is a notoriously ill-posed inverse problem. The received signal is a complex, entangled result of unknown geometry, unknown materials, and multi-path propagation effects.

To this end, we propose the Neural Electromagnetic Field (NEMF), a framework designed to systematically disentangle these coupled unknowns. Our key insight is that the ill-posed physics problem transforms into a well-posed one if the unknowns are resolved sequentially using multi-modal data. NEMF first reconstructs a high-fidelity geometric scaffold from images. This geometric prior then serves as a powerful constraint, enabling our framework to resolve the complex incident field ($f_\theta$). With both geometry and the incident field effectively constrained, the original problem is no longer ill-posed. This transformation unlocks our core contribution: a physics-supervised inversion module ($g_\phi$). This module is trained using a novel supervisory signal, computed by pairing the predicted incident field with a back-propagated measurement, to explicitly solve for the final material parameters ($\epsilon_r, \sigma$) via a differentiable physics layer.

We validate our framework on high-fidelity synthetic datasets, as this choice is essential for a rigorous quantitative validation against a known ground truth. Our experiments demonstrate that our joint optimization framework not only reconstructs three-dimensional EM parameter maps with high accuracy, but that the resulting functional digital twin can be used for high-fidelity \textbf{downstream simulation tasks} (e.g., channel prediction), significantly outperforming relevant baselines.

The main contributions of this paper are summarized as follows:
\begin{itemize}
    \item We propose NEMF, a novel multi-modal framework for dense, non-invasive physical inversion that builds functional, simulatable digital twins from visual and RF signals.
    \item We introduce a systematic disentanglement methodology where the image-based geometry serves as a powerful anchor; this enables the resolution of the ambient field, which in turn transforms the ill-posed problem into a well-posed, physics-supervised learning task.
    \item We experimentally validate this framework, demonstrating that our inversion accurately reconstructs continuous material parameter maps ($\epsilon_r, \sigma$) that enable high-fidelity physical simulation, significantly outperforming relevant baselines.
\end{itemize}
\section{Related Work}
\label{sec:related_work}

Our work is positioned at the intersection of three major fields: neural scene representation from computer vision, neural radiation fields from wireless communications, and traditional electromagnetic inversion from physics.

While Neural Radiance Fields (NeRF)~\cite{nerf} and its many variants for speed~\cite{muller2022instant, fridovich2022plenoxels, kerbl2023gaussians}, fidelity~\cite{barron2021mipnerf, verbin2022ref, nerfactor_2021}, and geometry~\cite{volsdf_2021, neus_2021, park2019deepsdf} excel at visual reconstruction, their implicit parameters are not physical. A rendered wall in a NeRF is functionally opaque to a physics simulator. Concurrent work has begun integrating physical properties like lighting and materials~\cite{zhang2021physg, nvdiffrec_2022} or system identification~\cite{phy_nerf_2023}. Our work leverages these geometric priors but extends the goal to EM physical function.

The "neural field" concept was recently extended to the RF domain~\cite{zhang2023signalnerf, muehlhaus2023radionerf, nerf2_2023}. Subsequent frameworks have formalized this for channel prediction~\cite{newrf_2024}, reconfigurable surfaces~\cite{r_nerf_2024}, generalizability~\cite{long2024gwrf}, and integration with 3DGS~\cite{carlson2024rf3dgs} or ray tracing~\cite{rf_ray_tracing_2025}. All these methods successfully adapt the NeRF architecture to solve the \textit{forward} problem: they interpolate the total radiation field from sparse $c_i^{obs}$ measurements. \textbf{However, a critical gap remains:} these black-box models cannot decouple the field into its constituents (incident field $\vec{E}_{inc}$, material properties $\Gamma$). Their goal is interpolation, not our goal of physical inversion.

In physics and engineering, traditional methods for inverting material parameters are well-established. These approaches, rooted in classical electrodynamics~\cite{jackson1999, hecht2017optics}, attempt to solve the inverse problem using rigorous physical models. Prominent examples include full-wave finite-element methods (FEM)~\cite{yee1966} and iterative optimization techniques such as Contrast Source Inversion (CSI)~\cite{icsi_2023, improved_csi_2022}. However, these methods, as summarized in foundational texts~\cite{colton2019inverse}, typically suffer from extreme non-convexity. They also require dense, calibrated measurements (a significant challenge in antenna theory~\cite{balanis2015antenna}), are highly sensitive to initialization and noise~\cite{haber2004} (often requiring strong regularization like total variation~\cite{rudin1992}), and are computationally prohibitive for large-scale, three-dimensional scenes. More recently, deep learning has been applied to these physics problems. This includes physics-informed neural networks (PINNs)~\cite{raissi2019}. This paradigm 
has been specifically applied to electromagnetic simulations \cite{pinn_em_field_2025} 
and wave propagation problems \cite{unified_neural_sim_2024}, as well as 
for inverse design~\cite{inverse_design_pinn_2025}, and other deep learning 
inversion techniques~\cite{inverse_dl_em_2022, unified_framework_2025}, including recent work on inverse rendering for mmWave signals~\cite{inverse_rendering_mmwave_2025}. While powerful, these methods often lack the robust geometric priors from vision and the powerful, coordinate-based representation our framework leverages.

Our work is also motivated by the emerging concept of 6G Network Digital Twins~\cite{6g_vision, network_dt_2025}, which envisions a future where wireless networks are mirrored by simulatable, physics-aware virtual replicas~\cite{towards_6g_dt_2023}. The goal is to create simulatable environments~\cite{boston_twin_2024} and high-fidelity RF maps~\cite{assessing_env_2025} using large channel models~\cite{large_model_dt_2024} to predict RF propagation. Unlike these works, which often rely on statistical or simplified ray-tracing~\cite{mathworks_raytrace}, our NEMF provides a method to \textbf{explicitly invert} the underlying physical material properties. This bridges the gap between visual scene capture and true, physics-aware functional simulation, enabling a new class of digital twin.
\section{Methodology}
\label{sec:method}

\begin{figure*}[t]
\centering
\includegraphics[width=\textwidth]{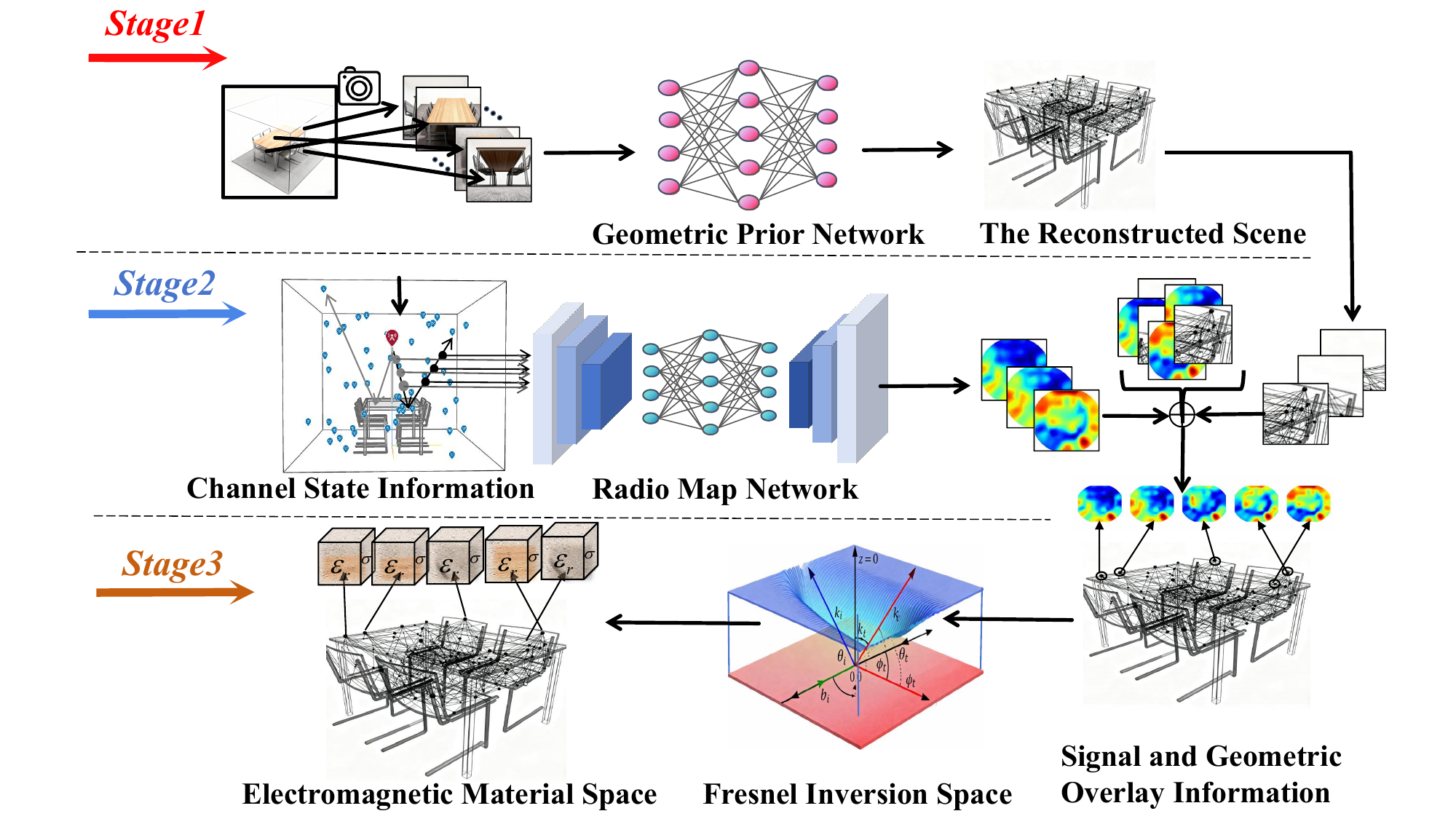} 
\caption{Our framework systematically decouples the electromagnetic inverse problem, which is challenging and ill posed, into three distinct stages. (1) Stage 1 reconstructs a geometric scaffold ($G$) with high fidelity. (2) Stage 2 reconstructs the ambient radio field ($f_\theta$). (3) Stage 3 uses the frozen $G$ and $f_\theta$ as priors to perform the final physics supervised inversion for the material properties ($g_\phi$).}
\label{fig:model_overview}
\end{figure*}

Reconstructing three dimensional electromagnetic parameter maps is an inverse problem that is highly challenging and ill posed. It is confounded by the coupled influence of unknown geometry, unknown incident fields, and unknown material properties.
To address this, we propose the Neural Electromagnetic Field (NEMF), an innovative framework that systematically decouples this problem into three distinct stages: (1) Geometric Prior Reconstruction, (2) Ambient Field Reconstruction, and (3) Physics-Supervised Material Inversion. As illustrated in Figure \ref{fig:model_overview}, we first isolate the geometry, then the field, which transforms the final material inversion from an ill posed problem into a well posed, supervised learning task. We detail each stage sequentially.

\subsection{Problem Definition}
The forward propagation of electromagnetic waves is governed by Maxwell's equations~\cite{jackson1999}, simplifying to the vector Helmholtz equation~\cite{jackson1999, balanis2015antenna} in the frequency domain:
\begin{equation}
\nabla^2 \mathbf{E} + k^2 \mathbf{E} = 0, \quad \text{where} \quad k^2 = \omega^2 \mu_0 \left(\epsilon_0 \epsilon_r - j \frac{\sigma}{\omega}\right)
\label{eq:helmholtz}
\end{equation}
Here, the complex wave number $k$ is directly modulated by the material parameters ($\epsilon_r, \sigma$). In our reflection model based on surfaces, the physics is governed by two stages. First, an incident vector field $\vec{E}_{inc}$ interacts with the surface, defined by a reflection operator $J$ (the Jones Matrix~\cite{born1999, balanis2015antenna}), to produce a reflected field $\vec{E}_{ref}$:
\begin{equation}
\vec{E}_{ref} = J \cdot \vec{E}_{inc}({x}, \omega)
\label{eq:e_ref_pred_full}
\end{equation}
Second, this reflected field propagates from the surface ${x}$ to a receiver at ${x}_{rx}$ via a propagation operator $\mathcal{P}$. This operator accounts for path loss from propagation in free space (FSPL, Eq. \ref{eq:pl_free}) and phase.
The final scalar measurement $c_i^{obs}$ is then captured by the receiver. This capture process is a dot product between the propagated field vector and the receiver antenna's fixed, known polarization vector, which we define as $\vec{h}_{rx}$ (e.g., $[1, 0]^T$ for a p polarized receiver). This yields the final Channel State Information (CSI):

\begin{equation}
\begin{split}
c_i^{obs} &= \mathcal{P}(\vec{E}_{ref}, d, \vec{h}_{rx}) \\
&= \vec{h}_{rx} \cdot \left( \vec{E}_{ref} \cdot 10^{-PL_{free} / 20} \cdot e^{-j (2\pi d f / c)} \right)
\end{split}
\label{eq:c_pred_final}
\end{equation}
\begin{equation}
PL_{free} = 20 \log_{10}(4\pi d f / c)
\label{eq:pl_free}
\end{equation}
The inverse problem is thus highly entangled: $c_i^{obs}$ depends on $G$ (which defines $d$ and $\theta_i$), $J$ (which defines the materials $\epsilon_r, \sigma$), and $\vec{E}_{inc}$, all of which are unknown. Our framework is designed to decouple these unknowns sequentially.

\subsection{Stage 1: Geometric Prior Reconstruction}
\label{sec:stage1_geo}
In the first stage, our objective is to decouple and fix the scene's geometry from images taken from multiple views. We employ the \textbf{instant-ngp-bound} framework~\cite{muller2022instant}, which trains a neural Signed Distance Function (SDF) with high fidelity. This network, parameterized by $f_{sdf}$, uses a hash grid with multiple resolutions to efficiently map a three dimensional spatial coordinate ${x}$ to its signed distance value $f_{sdf}({x})$. After training, we extract its zero level set $\mathcal{S} = \{{x} | f_{sdf}({x}) = 0\}$ as the surface and its gradient $\nabla f_{sdf}({x})$ as the surface normal $\mathbf{n}$. Together, these form a fixed geometric prior $G = (\mathcal{S}, \mathbf{n})$. This prior is crucial as it provides the precise surface boundaries, interaction point locations (${x}_{refl}$), and the geometric basis (e.g., the plane of incidence, incident angle $\theta_i$) required for the subsequent physical calculations.

\subsection{Stage 2: Ambient Field Reconstruction}
\label{sec:stage2a_field}
With the geometry $G$ frozen, the second stage is to reconstruct the ambient incident field, $\vec{E}_{inc}$. We represent this field using our Radio Map Network, $f_\theta$. This network is designed to learn a continuous representation of the incident field that varies spatially, implemented as a multilayer perceptron (MLP) based on NeWRF~\cite{newrf_2024}. It functions as a queryable network $f_\theta({x}, \mathbf{d})$ that predicts the complex electric field vector $\vec{E}_{inc}^{pred}$ arriving at any position ${x}$ from any direction $\mathbf{d}$.

To train $f_\theta$ in a logically consistent manner, we must supervise it with its intended target. Utilizing our synthetic dataset, we can compute the \textbf{ground truth incident electric field}, $\vec{E}_{inc}^{GT}({x}, \mathbf{d})$, at any 6D query point. We then train $f_\theta$ in a manner that is fully supervised to replicate this true incident field. During optimization, we minimize the L2 loss for a batch of sampled points $P$:
\begin{equation}
\mathcal{L}_{field} = \frac{1}{|P|} \sum_{({x}, \mathbf{d}) \in P} \| f_\theta({x}, \mathbf{d}) - \vec{E}_{inc}^{GT}({x}, \mathbf{d}) \|^2_2
\label{eq:field_loss}
\end{equation}
We also introduce a regularization term to ensure the spatial continuity of the predicted field:
\begin{equation}
\mathcal{L}_{reg} = \lambda \mathbb{E}_{({x}, \mathbf{d})} \left[ \| \nabla f_\theta({x}, \mathbf{d}) \|^2_2 \right]
\label{eq:reg_loss}
\end{equation}
The total loss is $\mathcal{L}_{total} = \mathcal{L}_{field} + \lambda_{reg}\mathcal{L}_{reg}$, where $\lambda_{reg}$ is a hyperparameter. Once this loss converges, the $f_\theta$ network is frozen. It now serves as a deterministic, queryable prior of the incident field for the final stage.
\label{sec:stage2b_material}
In the final stage, with both geometry $G$ and the field $f_\theta$ frozen, we solve for the material parameters using our decoder, $g_\phi$. This stage leverages the outputs from the previous two stages to form a supervised learning task that is well posed.

\subsection{Stage 3: Physics-Supervised Material Inversion}
\label{sec:stage2b_material}
In the final stage, with both geometry $G$ and the field $f_\theta$ frozen, we solve for the material parameters using our decoder, $g_\phi$. This stage leverages the outputs from the previous two stages to form a supervised learning task that is well posed.

The target label, $\mathbf{\Gamma}_{GT}$, is first computed. For a given reflection point ${x}$ and its corresponding measurement $c_i^{obs}$, we query our frozen Radio Map Network to get $\vec{E}_{inc}^{pred}$. Concurrently, we apply an operator $\mathcal{P}^{-1}$, which inverts the propagation model from propagation in free space, to the measurement $c_i^{obs}$ to compute the reflected field $\vec{E}_{ref}^{comp}$. With both fields known, we solve the linear system from Eq. \ref{eq:e_ref_pred_full} to compute the target Jones Matrix $\mathbf{\Gamma}_{GT}$ for that point and frequency.

Our decoder $g_\phi$ is not a simple MLP. It is a dedicated network architecture, designed to map a geometric point to its underlying physical properties, which are independent of frequency. 
Its input is the three dimensional coordinate ${x}$ of the point $P1$ provided by our geometric prior. This coordinate is processed by a sophisticated parallel encoding scheme. This scheme combines a classic sinusoidal basis function, which captures low frequency spatial context, with a high capacity, multi resolution hash grid. The hash grid is essential for this task, as it learns to map coordinates to a rich feature representation by interpolating entries from hash tables at several distinct levels of resolution. This design allows the network to efficiently capture fine geometric details and spatial variations. The features from both encoders are concatenated and passed to a deep residual MLP. This decoder then outputs a four dimensional latent vector $(a, b, c, d)$. A boundary layer using the Sigmoid function ensures these parameters are in a physically plausible range.

This latent vector models the physical frequency dispersion of the material using a power law relationship. Specifically, $a$ and $b$ model the dielectric constant $\epsilon_r$, while $c$ and $d$ model the conductivity $\sigma$, as a function of frequency $f$:
\begin{equation}
\hat{\epsilon}_r(f) = a \cdot f^b
\label{eq:epsilon_model}
\end{equation}
\begin{equation}
\hat{\sigma}(f) = c \cdot f^d
\label{eq:sigma_model}
\end{equation}
This latent vector $(a, b, c, d)$, along with the geometric angle $\theta_i$ and the specific query frequency $f_c$, is then parsed by our Differentiable Reflection Layer, $\mathcal{R}_{Frl}$. This layer, which is not trainable, performs the forward physics calculations. First, it computes the material properties for the given frequency: $\hat{\epsilon}_r(f_c)$ and $\hat{\sigma}(f_c)$ using Eq. \ref{eq:epsilon_model} and \ref{eq:sigma_model}. It then feeds these values into the differentiable Fresnel equations:
\begin{equation}
r_p = \frac{\eta_2 \cos \theta_i - \eta_1 \cos \theta_t}{\eta_2 \cos \theta_i + \eta_1 \cos \theta_t}
\label{eq:r_p}
\end{equation}
\begin{equation}
r_s = \frac{\eta_1 \cos \theta_i - \eta_2 \cos \theta_t}{\eta_1 \cos \theta_i + \eta_2 \cos \theta_t}
\label{eq:r_s}
\end{equation}
where the impedance $\eta$ is computed using the predicted properties that are dependent on frequency: $\eta = \sqrt{\mu / (\hat{\epsilon}_r(f_c) - j \hat{\sigma}(f_c) / (\omega \epsilon_0))}$. While our decoder learns $(a,b,c,d)$, it can also learn terms for cross polarization ($\hat{r}_{ps}, \hat{r}_{sp}$). The layer assembles the full Jones Matrix $\mathbf{\Gamma}_{pred}$, which is not diagonal:

\begin{equation}
\mathbf{\Gamma}_{pred}(f_c) = \mathcal{R}_{Frl}(g_\phi({x}), \theta_i, f_c) = \begin{pmatrix} r_p & \hat{r}_{ps} \\ \hat{r}_{sp} & r_s \end{pmatrix}
\label{eq:jones_matrix_full}
\end{equation}
\begin{figure}[t]
\centering
\includegraphics[width=\columnwidth]{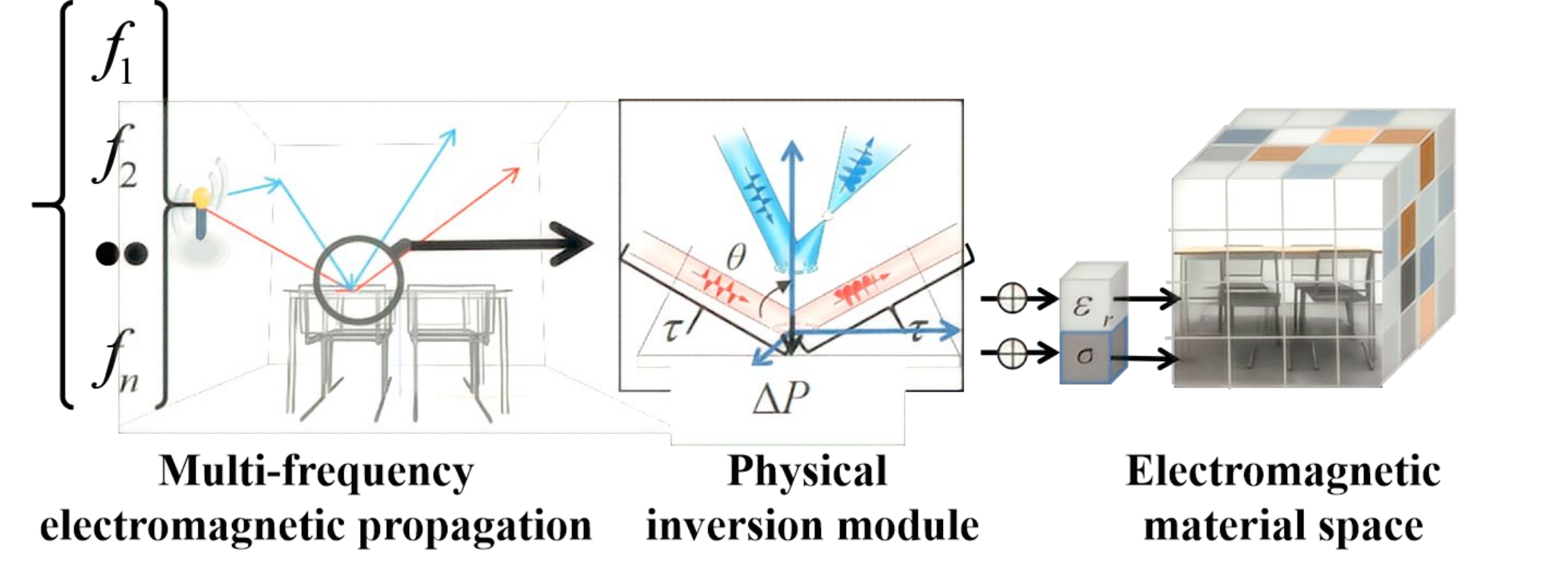} 
\caption{Fresnel Inversion Space. The complex relationship, which is not linear, between reflection and material parameters ($\epsilon_r, \sigma$) highlights the ambiguity of the inverse problem. Leveraging data from multiple frequencies robustly constrains the solution for our $\mathcal{L}_{total}$.}
\label{fig:fresnel_space}
\end{figure}

The final optimization loss is computed between this prediction $\mathbf{\Gamma}_{pred}(f_c)$ and our computed ground truth $\mathbf{\Gamma}_{GT}(f_c)$. Our total loss $\mathcal{L}_{total}$ combines the primary physics objective with regularization terms to ensure stable training and prevent overfitting of the hash grid, which has a high capacity. The primary loss $\mathcal{L}_{NMSE}$ is the \textbf{Normalized Mean Square Error}, summed over all frequencies $\omega$:
\begin{equation}
\mathcal{L}_{NMSE} = \sum_{\omega} \frac{\| \mathbf{\Gamma}_{GT} - \mathbf{\Gamma}_{pred} \|^2_F}{\| \mathbf{\Gamma}_{GT} \|^2_F}
\label{eq:new_loss}
\end{equation}
To control the complexity of the hash grid, we apply an L2 regularization $\mathcal{L}_{hash}$ on its embedding weights and an L1 sparsity regularization $\mathcal{L}_{gate}$ on its level gates. The final, combined loss function is:
\begin{equation}
\mathcal{L}_{total} = \mathcal{L}_{NMSE} + \lambda_{hash}\mathcal{L}_{hash} + \lambda_{gate}\mathcal{L}_{gate}
\label{eq:full_loss}
\end{equation}
where $\lambda_{hash}$ and $\lambda_{gate}$ are hyperparameters that balance fidelity and regularization. By optimizing this loss, the gradient flows through the differentiable reflection layer and backpropagates through the residual MLP and the hash grid. This supervised process forces the network $g_\phi$ to find the interpretable parameters $(a,b,c,d)$, which are independent of frequency and grounded in physics, that best explain the observed reflection behavior.

\section{Experiments}
\label{sec:experiments}

We conduct a comprehensive set of experiments to rigorously validate our NEMF framework. We first establish its significant quantitative advantage over "black box" MLP baselines in material inversion accuracy. We then present a systematic ablation study to dissect how specific architectural choices within our hash grid framework quantitatively impact performance across multiple scenes. Finally, we explore advanced optimization strategies, such as LBFGS finetuning and weighted losses, to further refine the physical accuracy of the inversion.

% --- 4.1. Experimental Setup ---
\subsection{Experimental Setup}
\label{sec:setup}

\paragraph{Geometric Prior (Stage 1).}
We first use the \textbf{instant-ngp-bound} framework to train a Signed Distance Function (SDF) model with high fidelity from 100 images taken from multiple views for each scene. This provides the fixed geometric prior $G$ (surface points $P1$, normals) with high resolution required for our Stage 2 physical inversion.

\begin{figure*}[t]
\centering
\subfloat[Office Scene]{
    \includegraphics[width=0.31\textwidth]{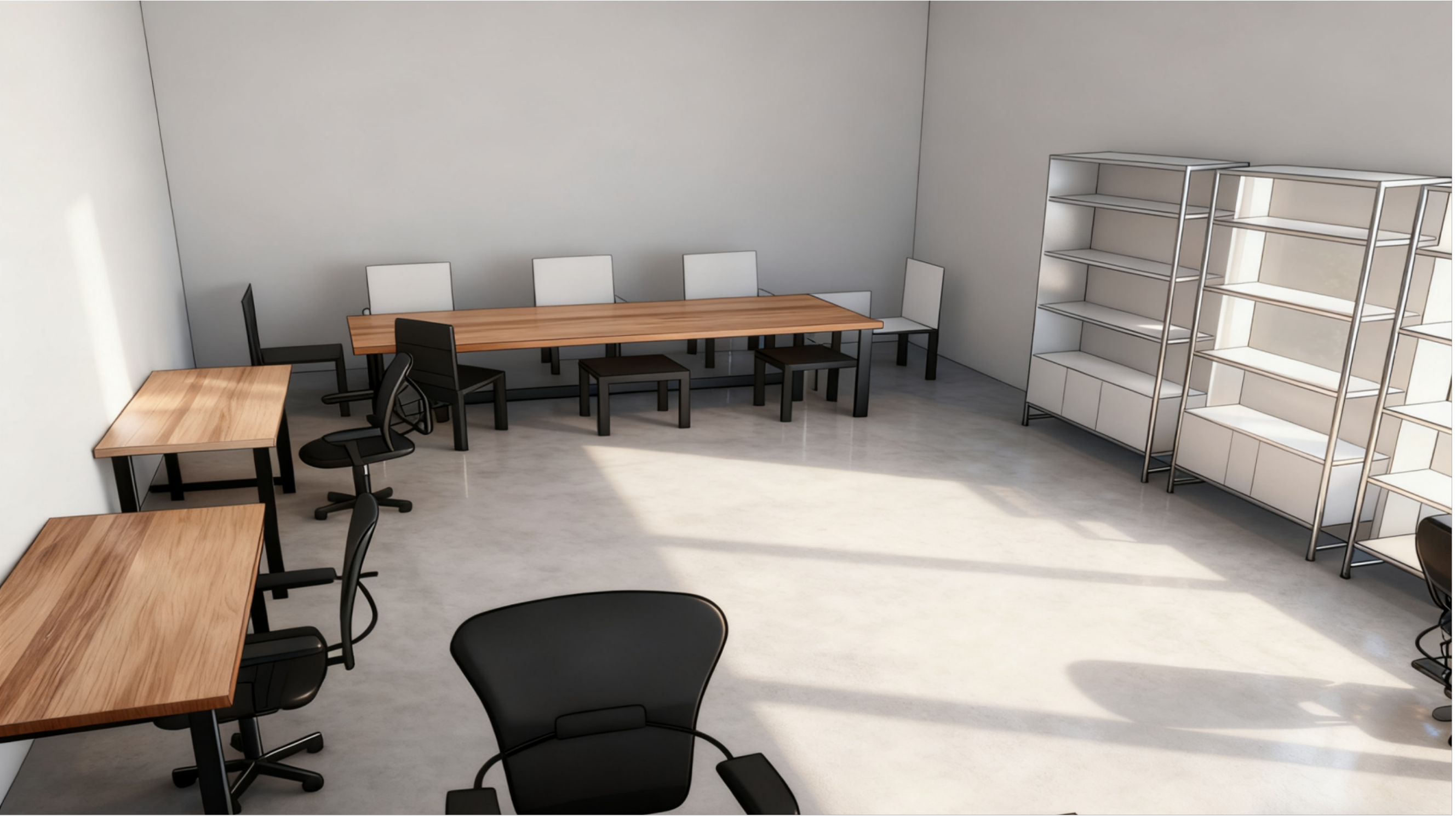}
    \label{fig:scene_office}
}
\hfill
\subfloat[Bedroom Scene]{
    \includegraphics[width=0.31\textwidth]{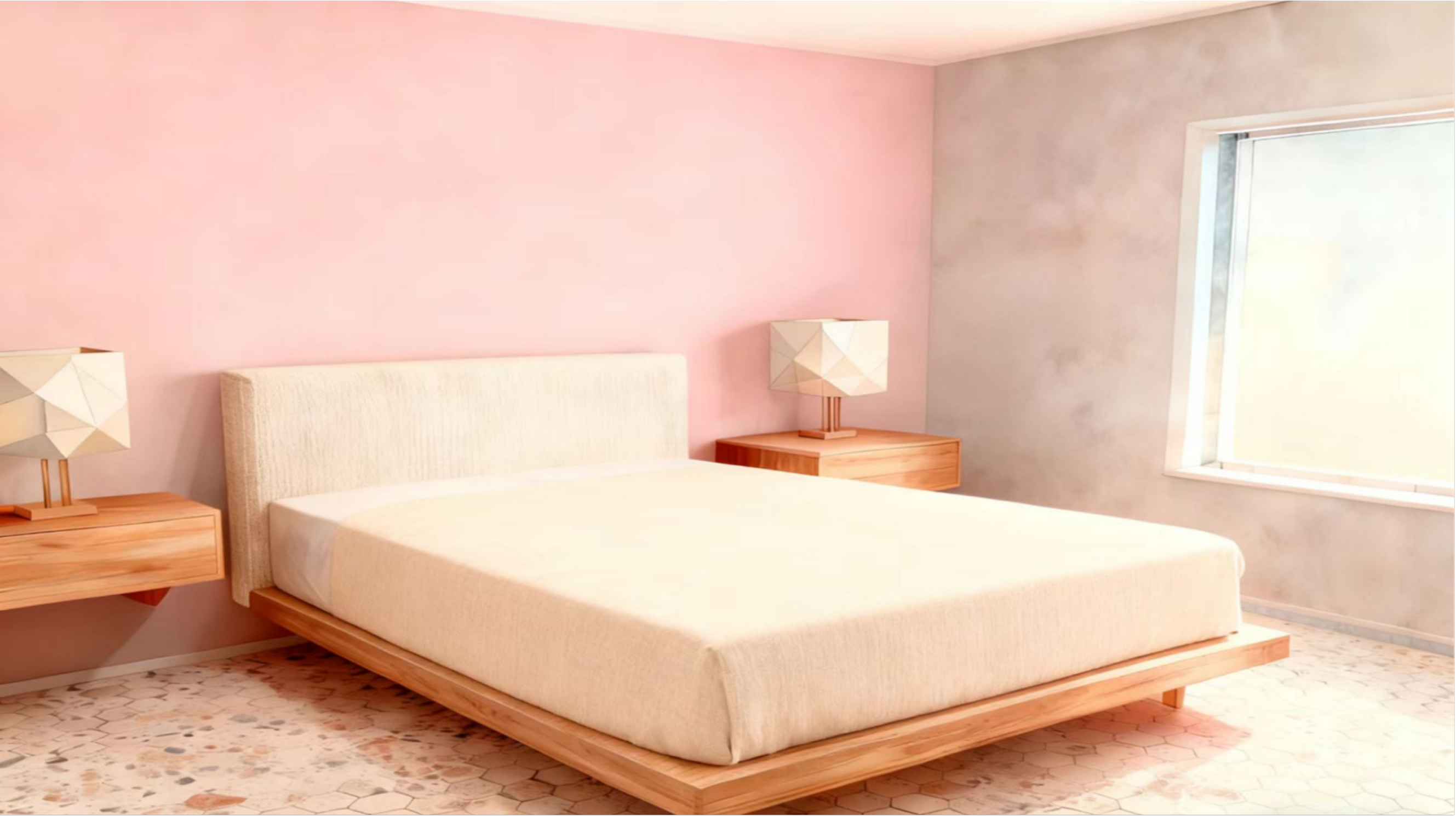}
    \label{fig:scene_bedroom}
}
\hfill
\subfloat[Conference Room]{
    \includegraphics[width=0.31\textwidth]{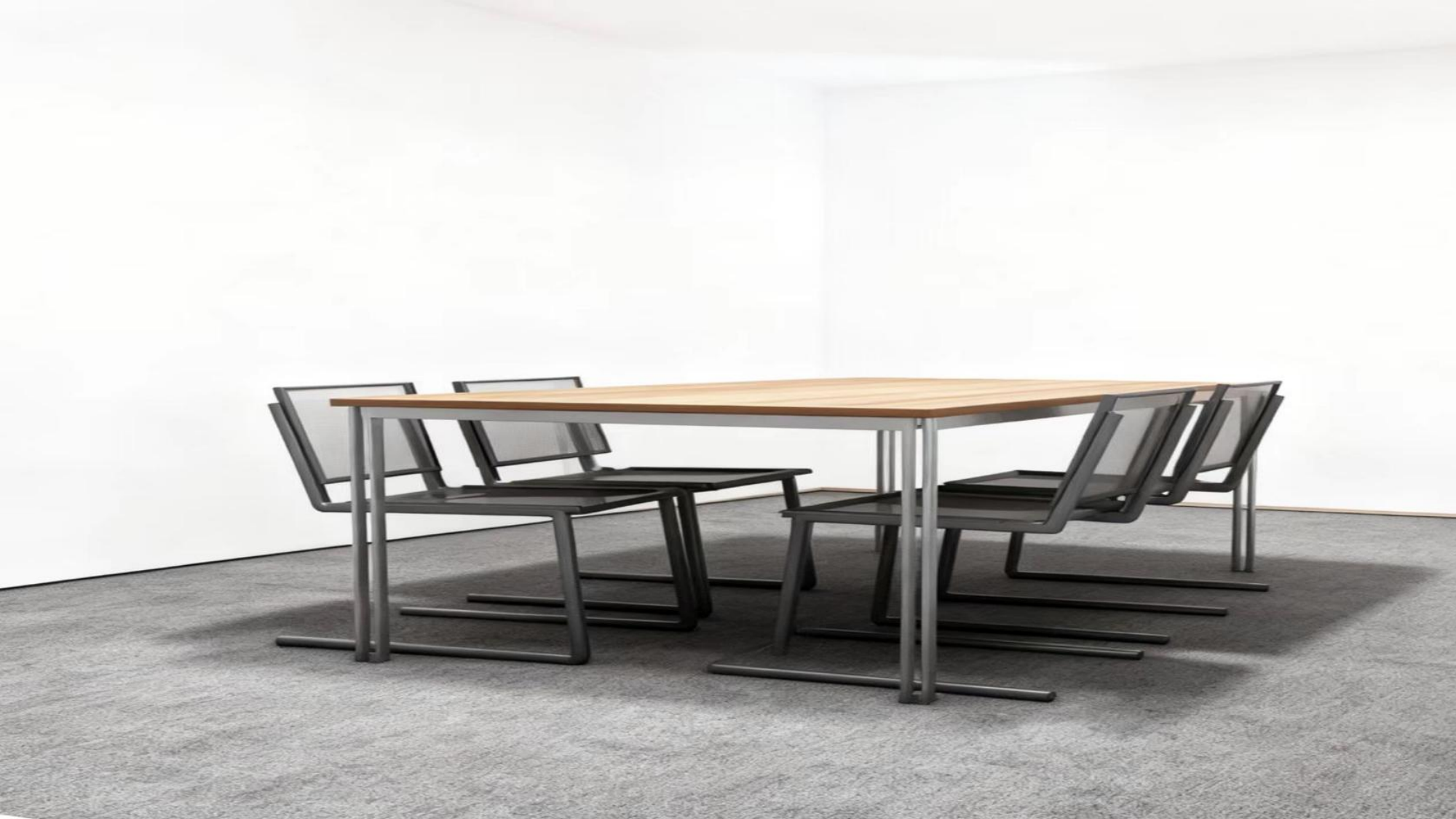}
    \label{fig:scene_confroom}
}
\caption{The three indoor scenes used in our synthetic dataset. We use these geometric models, which possess high fidelity, to generate both the images from multiple views for Stage 1 and the sparse CSI data from ray tracing for Stage 2.}
\label{fig:scenes}
\end{figure*}

\paragraph{Datasets.}
Our physical inversion experiments are conducted on synthetic datasets with high fidelity for three indoor scenes: \textit{Office}, \textit{Bedroom}, and \textit{Conference Room}, which are visualized in Figure \ref{fig:scenes}. Each scene possesses an approximate volume of $5\text{m} \times 4\text{m} \times 3\text{m}$ and incorporates diverse physical materials, including concrete, wood, and glass, providing varied material parameters ($\epsilon_r$ and $\sigma$). The data generation utilizes a simulator based on MATLAB for rigorous ray tracing, accounting for multiple reflections and complex diffraction effects at high frequencies. This process generates sparse CSI measurements ($G_{\text{true}}$) at only 500 random receiver (RX) locations from one fixed transmitter (TX). The resulting dataset is characterized by high sparsity, which accurately reflects the challenge of practical, non-invasive measurement scenarios. Data is consistently sampled at 8 discrete frequencies (2.4--5.8 GHz).

% --- [场景图] ---
% --- 场景图结束 ---

\paragraph{Evaluation Metrics.}
While our model is trained on a Complex MSE loss to facilitate gradient propagation, effective evaluation requires physical metrics that are readily interpretable by engineers. We employ a hierarchy of metrics. The first level targets the latent space, where we report the Mean Absolute Error (MAE) of the intermediate physical parameters (\textbf{ABCD-MAE}). The final and most critical metrics quantify the decoded material properties: \textbf{Dielectric Constant ($\epsilon_r$ MRE)} and \textbf{Conductivity ($\sigma$ MRE)}. The Mean Relative Error (MRE) is used for these final metrics because the target values for $\epsilon_r$ and $\sigma$ span multiple orders of magnitude across diverse materials, making relative comparison essential for a fair assessment of physical fidelity.

\paragraph{Implementation Details.}
All models are implemented in PyTorch~\cite{paszke2019pytorch} and trained on a single NVIDIA RTX 4090 GPU. We use the Adam optimizer~\cite{kingma2014adam} with a batch size of 32768. Unless otherwise specified (in ablations), our NEMF models use the \textbf{Arch-0} configuration (see Table \ref{tab:ablation_architecture}) and are trained with a learning rate of $1.0 \times 10^{-3}$ for the MLP and $5.0 \times 10^{-4}$ for the hash grid, scheduled by \texttt{ReduceLROnPlateau}.

% --- 4.2. Main Results: Framework Comparison ---
\subsection{Main Results: Framework Comparison}
\label{sec:main_results}

We first compare our NEMF framework against the "black box" MLP baselines.

\paragraph{Baselines (MLPs).}
The MLP baselines represent the standard neural field approach. They take the concatenated vector of coordinates and frequency, passed through positional encoding, as input. A single MLP is then used to directly predict the complex response $G_{\text{pred}}$. This "entangled" architecture is forced to simultaneously learn spatial geometry and physics dependent on frequency with a single set of weights. We test multiple configurations and report the strongest performing variant, which is an eight layer, eight frequency model.

\paragraph{Our Model (NEMF).}
As detailed in Section \ref{sec:method}, our model decouples this problem. A powerful hash grid learns the spatial features $f(\mathbf{x})$, which an MLP then maps to physical parameters $(a,b,c,d)$ that are independent of frequency. A final layer that is not trainable then computes the response $G_{\text{pred}} = \mathcal{P}(a,b,c,d, f_c)$, which is dependent on frequency.

\paragraph{Quantitative Comparison.}
\textbf{Table \ref{tab:main_results}} presents the quantitative comparison, broken down by scene. The results are conclusive. Our NEMF framework (using the strong \textbf{Opt-B} configuration) significantly outperforms the strongest MLP baseline across all key physical metrics. While the baseline achieves a reasonable average \textit{Eps MRE} of 0.078, our decoupled framework reduces this error by over seven times to 0.011. The improvement in \textit{Sigma MRE} is also substantial, with our model more than halving the error from 0.639 to 0.317. This confirms our hypothesis: the entangled "black box" architecture is fundamentally unsuited for this problem, while our physics-decoupled framework achieves a dramatic and consistent improvement in accuracy.

% ===== TABLE 1: MAIN RESULTS (Per-Scene Breakdown) - FINAL WIDE TABLE =====
\begin{table*}[t]
\centering
\caption{Main quantitative comparison against the strongest MLP baseline, broken down by scene. Our decoupled NEMF framework demonstrates a significant and consistent reduction in material inversion error across all test environments.}
\label{tab:main_results}
\small 
\begin{tabular*}{\textwidth}{l @{\extracolsep{\fill}} | c c | c c | c c}
\hline
 & \multicolumn{2}{c |}{\textbf{Conference Room}} & \multicolumn{2}{c |}{\textbf{Bedroom}} & \multicolumn{2}{c}{\textbf{Office}} \\
\textbf{Model} & \textbf{Eps MRE} $\downarrow$ & \textbf{Sigma MRE} $\downarrow$ & \textbf{Eps MRE} $\downarrow$ & \textbf{Sigma MRE} $\downarrow$ & \textbf{Eps MRE} $\downarrow$ & \textbf{Sigma MRE} $\downarrow$ \\
\hline
MLP Baseline (8-Layer, 8-Freq) & 0.071 & 0.497 & 0.102 & 0.774 & 0.062 & 0.647 \\
\textbf{Ours (NEMF, Opt-B)} & \textbf{0.021} & \textbf{0.164} & \textbf{0.004} & \textbf{0.114} & \textbf{0.007} & \textbf{0.672} \\
\hline
\end{tabular*}
\end{table*}
% ========================================================

% --- 4.3. In-Depth Ablation Study ---
\subsection{In-Depth Ablation Study}
\label{sec:ablations}

Having confirmed our framework's macro-level superiority, we conduct a systematic ablation study to dissect the internal components of NEMF.

\paragraph{Architectural Components (Arch-0 to Arch-7).}
We first analyze the impact of "static" architectural choices. We establish a medium capacity baseline (\textbf{Arch-0}) and systematically vary a single component at a time. The results across all three scenes are detailed in \textbf{Table \ref{tab:ablation_architecture}}. 
Analysis of this table reveals several nuanced insights that are remarkably consistent across all scenes. First, the impact of \textbf{PE frequency} (\textbf{Arch-1}) is mixed, with the \textit{Office} and \textit{Bedroom} scenes showing a slight improvement in \textit{Eps MRE}, while the \textit{Conference Room} degrades. This suggests that the hash grid is the dominant feature extractor, making the model less sensitive to PE bandwidth. \textbf{Crucially, the capacity ablations also reflect the quality of the geometric priors.} Second, we explored \textbf{hash capacity}. Increasing capacity, particularly via extreme combinations like (\textbf{Arch-6}), yields significant gains in the more complex \textit{Conference Room} (0.013 $\rightarrow$ 0.006) and \textit{Bedroom} (0.029 $\rightarrow$ 0.011) scenes. This highlights a clear trade-off between model capacity and scene complexity, as better hash encoding allows for a more accurate extraction of surface normals. Finally, we validated our \textbf{key mechanisms}. Disabling either the gating mechanism (\textbf{Arch-7}) or the skip connection (not shown) consistently causes performance degradation, confirming their necessity.

% ===== TABLE 2: ARCHITECTURE ABLATION (B0-B7) - 3 SCENES - DATA-FILLED =====
\begin{table*}[!t]
\centering
\caption{Ablation study on NEMF architectural components. We report key physical metrics across all three scenes. The baseline (\textbf{Arch-0}) is highlighted. Disabling core components (Arch-1, 7, 8) consistently results in severe performance degradation.}
\label{tab:ablation_architecture}
\resizebox{\textwidth}{!}{%
\begin{tabular}{l l l | c c | c c | c c}
\hline
& & & \multicolumn{2}{c |}{\textbf{Conference Room}} & \multicolumn{2}{c |}{\textbf{Bedroom}} & \multicolumn{2}{c}{\textbf{Office}} \\
\textbf{ID} & \textbf{Component} & \textbf{Configuration} & \textbf{Eps MRE} $\downarrow$ & \textbf{Sigma MRE} $\downarrow$ & \textbf{Eps MRE} $\downarrow$ & \textbf{Sigma MRE} $\downarrow$ & \textbf{Eps MRE} $\downarrow$ & \textbf{Sigma MRE} $\downarrow$ \\
\hline
\textbf{Arch-0} & \textbf{Baseline Config.} & \textbf{PE L=16, L=10, res=24, feat=8, gate=on} & \textbf{0.013} & \textbf{0.447} & \textbf{0.029} & \textbf{0.837} & \textbf{0.013} & \textbf{0.710} \\
\hline
Arch-1 & PE Frequency & PE L=10 & 0.015 & 0.552 & 0.027 & 0.816 & 0.008 & 0.705 \\
Arch-2 & Hash Layers & L=12 & 0.007 & 0.353 & 0.025 & 0.818 & 0.015 & 0.690 \\
Arch-3 & Hash Resolution & base\_res=32 & 0.012 & 0.467 & 0.027 & 0.811 & 0.014 & 0.708 \\
Arch-4 & Arch-2 + Arch-3 & L=12, res=32 & 0.012 & 0.556 & 0.015 & 0.399 & 0.016 & 0.704 \\
Arch-5 & Hash Features & feat=16 & 0.011 & 0.509 & 0.023 & 0.806 & 0.007 & 0.634 \\
Arch-6 & Arch-2 + Arch-5 & L=12, feat=16 & \textbf{0.006} & \textbf{0.335} & \textbf{0.011} & \textbf{0.367} & 0.017 & 0.697 \\
Arch-7 & Gating & gate=off & 0.017 & 0.569 & 0.027 & 0.809 & 0.015 & 0.705 \\
\hline
\end{tabular}%
}
\end{table*}
% ==================================================

\paragraph{Optimization Strategies (Opt-A to Opt-F).}
We next analyze "dynamic" training strategies, with results across all scenes summarized in \textbf{Table \ref{tab:ablation_optimization}}. These results provide critical insights. First and most importantly, \textbf{LBFGS finetuning} (\textbf{Opt-B}) provides a dramatic and consistent improvement over the Adam-only baseline (\textbf{Opt-A}). It reduces the \textit{Eps MRE} by $\sim$42\% in the \textit{Conference Room}, a staggering $\sim$83\% in the \textit{Bedroom}, and $\sim$59\% in the \textit{Office}. This demonstrates that our loss landscape, while complex, benefits immensely from optimization of the second order. Second, by comparing (\textbf{Opt-B}) to (\textbf{Opt-F}), we see that finetuning the \textbf{MLP only} provides almost no benefit, or even degrades performance. This proves the LBFGS gains come from finetuning the \textit{entire} network, particularly the hash grid embeddings. Third, \textbf{Extreme Capacity} (\textbf{Opt-C}) shows mixed results, underperforming the standard \textbf{Opt-B} in the \textit{Bedroom} scene, suggesting \textbf{Opt-B} hits a sweet spot of capacity and optimization. Finally, the \textbf{weighted loss} strategy (\textbf{Opt-E}) did not yield the expected gains, performing worse than the \textbf{Opt-B} baseline, suggesting this is a sensitive tradeoff.

% ===== TABLE 3: OPTIMIZATION ABLATION (A-F) - 3 SCENES - DATA-FILLED =====
% --- 逻辑归位：此表格应在 4.3 节的 Opt 分析段之后 ---
\begin{table*}[!t]
\centering
\caption{Ablation study on optimization strategies. We report the final \textit{Eps ($\epsilon_r$) MRE} and \textit{Sigma ($\sigma$) MRE} across all scenes.}
\label{tab:ablation_optimization}
\resizebox{\textwidth}{!}{%
\begin{tabular}{l l l | c c | c c | c c}
\hline
& & & \multicolumn{2}{c |}{\textbf{Conference Room}} & \multicolumn{2}{c |}{\textbf{Bedroom}} & \multicolumn{2}{c}{\textbf{Office}} \\
\textbf{ID} & \textbf{Configuration} & \textbf{Strategy} & \textbf{Eps MRE} $\downarrow$ & \textbf{Sigma MRE} $\downarrow$ & \textbf{Eps MRE} $\downarrow$ & \textbf{Sigma MRE} $\downarrow$ & \textbf{Eps MRE} $\downarrow$ & \textbf{Sigma MRE} $\downarrow$ \\
\hline
Opt-A & Medium Capacity & Baseline (W256/D6) & 0.036 & 0.334 & 0.024 & 0.763 & 0.017 & 0.713 \\
\textbf{Opt-B} & \textbf{High Capacity} & \textbf{+ LBFGS Finetuning} & \textbf{0.021} & \textbf{0.164} & \textbf{0.004} & \textbf{0.114} & \textbf{0.007} & \textbf{0.672} \\
Opt-C & Extreme Capacity & + Progressive Unfreeze & 0.021 & 0.145 & 0.025 & 0.847 & 0.015 & 0.680 \\
Opt-D & Medium Capacity & + Progressive Unfreeze & 0.046 & 0.352 & 0.023 & 0.773 & 0.013 & 0.708 \\
Opt-E & High Capacity (B) & + High-Freq Loss Weight & 0.020 & 0.123 & 0.037 & 0.348 & 0.015 & 0.681 \\
Opt-F & Medium Capacity & + LBFGS (MLP Only) & 0.035 & 0.331 & 0.028 & 0.781 & 0.015 & 0.708 \\
\hline
\end{tabular}%
}
\end{table*}

\subsection{Qualitative Analysis and Limitations}
\label{sec:qualitative_limitations}

\paragraph{Qualitative Results.}
Our qualitative analysis provides definitive visual validation of the quantitative results. We observe that the MLP baseline fails completely. Its predictions are noisy, spatially non-uniform, and bear no correspondence to the underlying geometry. This proves its inability to learn the physics geometry coupling. In sharp contrast, our NEMF framework (using the Opt-B configuration) reconstructs material maps that are spatially coherent, possess sharp boundaries at material interfaces, and are physically accurate. These reconstructed maps closely match the ground truth. Crucially, the residual prediction error in our model is not random. It is localized almost exclusively to complex geometric boundaries and sharp corners. This visual evidence confirms that the primary source of residual error stems from the geometric instability of the local normal vector ($\mathbf{n}$) derivation in these difficult zones, rather than a failure of the core material inversion logic. This visual confirmation reinforces the sensitivity to normal vector accuracy demonstrated in our ablation study.

\paragraph{Limitations.}
Despite the strong results, our framework has limitations. First, the performance of our Stage 2 inversion is highly dependent on the accuracy of the geometric prior from Stage 1. Any error in the derived normal vectors ($\mathbf{n}$) or the angle of incidence ($\theta_i$) propagates non-linearly through the Fresnel equations, leading to localized prediction failure. Second, our validation is currently based on synthetic data with high fidelity. Real-world CSI measurements will introduce more complex noise profiles, hardware calibration errors, and more severe multipath effects, which we leave for future work.
\section{Conclusion}
\label{sec:conclusion}

In this paper, we introduced the Neural Electromagnetic Field (NEMF), a novel framework using multiple modalities for reconstructing functional digital twins. We invert electromagnetic material properties from non-invasive visual and RF signals. We show that by systematically disentangling the inverse problem, which is ill posed, we anchor the process with geometry that has high fidelity to resolve the ambient field. This transforms the task into a well posed learning problem supervised by physics. Our core contributions, including the sequential pipeline for decoupling and the layer for reflection that is differentiable, enable accurate recovery of continuous material maps ($\epsilon_r, \sigma$).

Experiments on diverse synthetic scenes demonstrate that NEMF significantly outperforms baselines in material inversion accuracy, with reductions in error of an order of magnitude. The resulting functional twins support simulations of the downstream type with high fidelity, bridging the gap between visual appearance and physical interaction.

While our framework advances physical modeling that is not invasive, it relies on synthetic data for validation. Future work will extend to measurements in the real world, incorporating robustness to noise and integration with emerging 6G sensing protocols. This opens new avenues for applications in robotics, AR/VR, and wireless optimization, paving the way for truly simulatable digital worlds.

{
    \small
    \bibliographystyle{ieeenat_fullname}
    \bibliography{main}
}


\end{document}